\documentclass{article} % For LaTeX2e
\usepackage{iclr2018_conference,times}
\usepackage{hyperref}
\usepackage{varioref}
\usepackage{url}
\def\*#1{\mathbf{#1}}
\usepackage{multirow}
\usepackage{tabularx}
\usepackage[draft]{todonotes}
\usepackage{xcolor}
\usepackage{subfigure}
 
\usepackage{color}
\newcommand{\samira}[1]{\textcolor{blue}{#1}}

\usepackage{algorithm}
\usepackage{algorithmic}
\usepackage[fleqn]{amsmath}
\usepackage{amsthm}
\usepackage{amssymb}

\title{Variational Bi-LSTMs}

\author{Samira Shabanian$^{1,}$\thanks{samira.shabanian@microsoft.com}\ ,    Devansh Arpit$^2$, Adam Trischler$^3$ \& Yoshua Bengio$^{4}$  \\
$^{1,3}$ Microsoft Maluuba\\
$^{2,4}$ MILA, Universit\'e de Montr\'eal\\
}

\iclrfinalcopy % Uncomment for camera-ready version, but NOT for submission.

\begin{document}

\maketitle

\begin{abstract}
Recurrent neural networks like long short-term memory (LSTM) are important architectures for sequential prediction tasks. LSTMs (and RNNs in general) model sequences along the forward time direction. Bidirectional LSTMs (Bi-LSTMs) on the other hand model sequences along both forward and backward directions and are generally known to perform better at such tasks because they capture a richer representation of the data. In the training of Bi-LSTMs, the forward and backward paths are learned independently. We propose a variant of the Bi-LSTM architecture, which we call Variational Bi-LSTM, that creates a channel between the two paths (during training, but which may be omitted during inference); thus optimizing the two paths jointly. We arrive at this joint objective for our model by minimizing a variational lower bound of the joint likelihood of the data sequence. Our model acts as a regularizer and encourages the two networks to inform each other in making their respective predictions using distinct information. We perform ablation studies to better understand the different components of our model and evaluate the method on various benchmarks, showing state-of-the-art performance.
\end{abstract}

\section{Introduction}
Recurrent neural networks (RNNs) have become the standard models for sequential prediction tasks with state of the art performance in a number of applications like sequence prediction, language translation, machine comprehension, and speech synthesis~\citep{arik2017deep, wang2017tacotron, mehri2016samplernn, sotelo2017char2wav}.
RNNs model temporal data by encoding a given arbitrary-length input sequentially, at each time step combining a transformation of the current input with the encoding from the previous time step. This encoding, referred to as the RNN hidden state, summarizes all previous input tokens.

Viewed as ``unrolled'' feedforward networks, RNNs can become arbitrarily deep depending on the input sequence length, and use a repeating module to combine the input with the previous state at each time step. Consequently, they suffer from the vanishing/exploding gradient problem~\citep{DBLP:journals/corr/abs-1211-5063}.
This problem has been addressed through architectural variants like the long short-term memory (LSTM)~\citep{lstm} and the gated recurrent unit (GRU)~\citep{gru}. These architectures add a linear path along the temporal sequence which allows gradients to flow more smoothly back through time.

Various regularization techniques have also been explored to improve RNN performance and generalization.
Dropout \citep{srivastava2014dropout} regularizes a network by randomly dropping hidden units during training. However, it has been observed that using dropout directly on RNNs is not as effective as in the case of feed-forward networks. To combat this, \citet{zaremba2014recurrent} propose to instead apply dropout on the activations that are not involved in the recurrent connections (Eg. in a multi-layer RNN); \cite{vdropout} propose to apply the same dropout mask through an input sequence during training. In a similar spirit to dropout, Zoneout \citep{DBLP:journals/corr/KruegerMKPBKGBL16} proposes to choose randomly whether to use the previous RNN hidden state.

The aforementioned architectures model sequences along the forward direction of the input sequence. Bidirectional-LSTM, on the other hand, is a variant of LSTM that simultaneously models each sequence in both the forward and backward direction. This enables a richer representation of data, since each token's encoding contains context information from the past and the future.
It has been shown empirically that bidirectional architectures generally outperform unidirectional ones on many sequence-prediction tasks. However, the forward and backward paths in Bi-LSTMs are trained separately and the benefit usually comes from the combined hidden representation from both paths. In this paper, our main idea is to frame a joint objective for Bi-directional LSTMs by minimizing a variational lower bound of the joint likelihood of the training data sequence. This in effect implies using a variational auto-encoder (VAE; \citet{vae}) that takes as input the hidden states from the two paths of the Bi-LSTM and maps them to a shared hidden representation of the VAE at each time step. The samples from the VAE's hidden state are then used for reconstructing the hidden states of both the LSTMs. While the use of a shared hidden state acts as a regularizaer during training, the dependence on the backward path can be ignored during inference by sampling from the VAE prior. Thus our model is applicable in domains where the future information is not available during inference (Eg. language generation). We refer to our model as Variational Bi-LSTM. We note that recently proposed methods like TwinNet \citep{serdyuk2017twin} and Z-forcing \citep{Sordoni2017} are similar in spirit to this idea. We discuss the differences between our approach and these models in section \ref{sec_related_work}.

% In this paper, our main idea is to create a channel of information between the two paths that acts as a regularization during training, but doesn't hinder inference even in the absence of the other path (the backward path in practical scenarios). 

% We note that recently proposed methods like TwinNet \cite{serdyuk2017twin} and Z-forcing \cite{Sordoni2017} are similar in spirit to this idea.
% In our approach, we use a variational auto-encoder (VAE; \citet{vae}) that takes as input the hidden states from the two paths of the Bi-LSTM and maps them to a shared hidden representation of the VAE at each time step. The samples from the VAE hidden state are then used both for reconstructing the LSTM hidden states and feeding forward to the next hidden state. In this way, we create a \textit{channel} between the two paths that acts as a regularization for learning better representations. We refer to the resulting model as a Variational Bi-LSTM.

%HERE YOU SHOULD SUMMARIZE THE CONTRIBUTIONS AND RESULTS OF THE PAPER
%\samira{We first introduce Variational Bi-LSTMs in full details and then the ability to model complex distributions of our approach are shown empirically.}
%\smra{@Devansh @Adam}
Below, we describe Variational Bi-LSTMs in detail and then demonstrate empirically their ability to model complex sequential distributions. In experiments, we obtain state-of-the-art or competitive performance on the tasks of Penn Treebank, IMDB, TIMIT, Blizzard, and Sequential MNIST. %\smra{@Adam NICE}

\section{Variational Bi-LSTM}\label{sec:md}
Bi-LSTM is a powerful architecture for sequential tasks because it models temporal data both in the forward and backward direction. For this, it uses two LSTMs that are generally learned independent of each other; the richer representation in Bi-LSTMs results from combining the hidden states of the two LSTMs, where combination is often by concatenation. The idea behind variational Bi-LSTMs is to create a \textit{channel} of information exchange between the two LSTMs that helps the model to learn better representations. We create this dependence by using the variational auto-encoder (VAE) framework. This enables us to take advantage of the fact that VAE allows for sampling from the prior during inference. For sequence prediction tasks like language generation, while one can use Bi-LSTMs during training, there is no straightforward way to employ the full bidirectional model during inference because it would involve, Eg., generating a sentence starting at both its beginning and end. In such cases, the VAE framework allows us to sample from the prior at inference time to make up for the absence of the backward LSTM.

% are generative autoregressive networks that can learn to generate backward LSTM path  and benefit from it during inference. Here we describe Variational Bi-LSTMs in full detail and how it can be trained to become more strong. %using two auxiliary costs and skip gradient trick. % This way and have a better prediction.

Now we describe our variational Bi-LSTM model formally. Let $\*X=\{\*x^{(i)}\}_{i=1}^N$ be a dataset consisting of $N$ i.i.d. sequential data samples of continuous or discrete variables. For notational convenience, we will henceforth drop the superscript $i$ indexing samples. For each sample sequence $\*x=(\*x_1, \ldots, \*x_T)$, the hidden state of the forward LSTM is given by:
\begin{align}
\*h_{t} = \overrightarrow{f}(\*x_{t}, \*h_{t-1}, \*z_{t}, {\tilde{\*b}}_{t}).
\end{align} 
The hidden state of the backward LSTM is given by,
 \begin{align}
\*b_{t} = \overleftarrow{f}(\*x_{t}, \*b_{t + 1}).
\end{align}
In the forward LSTM, $\overrightarrow{f}$ represents the standard LSTM function modified to account for the additional arguments used in our model using separate additional dense matrices for $ \*z_{t}$ and $ {\tilde{\*b}}_{t}$. The function $\overleftarrow{f}$ in the backward LSTM is defined as in a standard Bi-LSTM.

\begin{figure}
  \centering
    \subfigure[Training phase of variational Bi-LSTM]{
    \label{fig:tr}
    \includegraphics[scale=0.5]{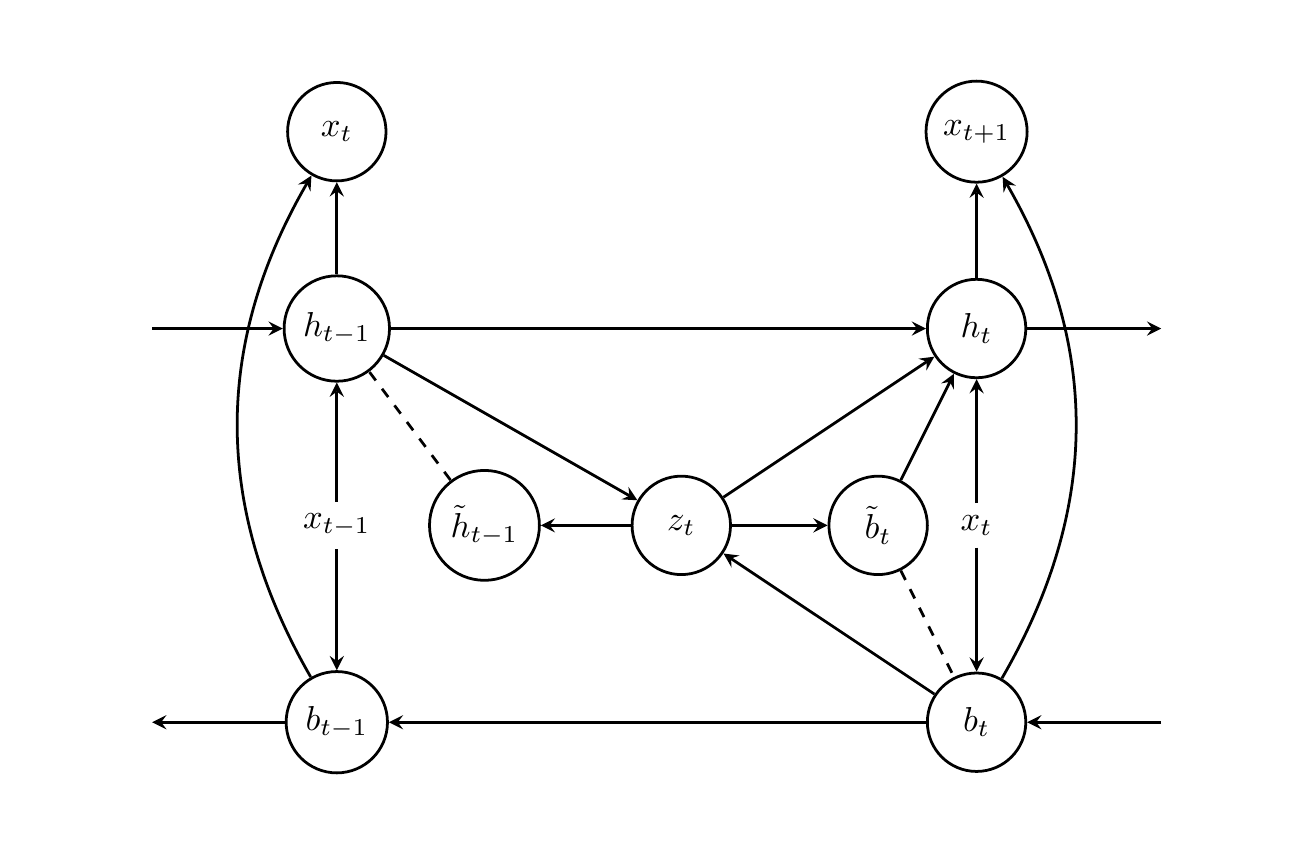}
  }
  \subfigure[Inference phase of variational Bi-LSTM]{
    \label{fig:inf}
    \includegraphics[scale=0.5]{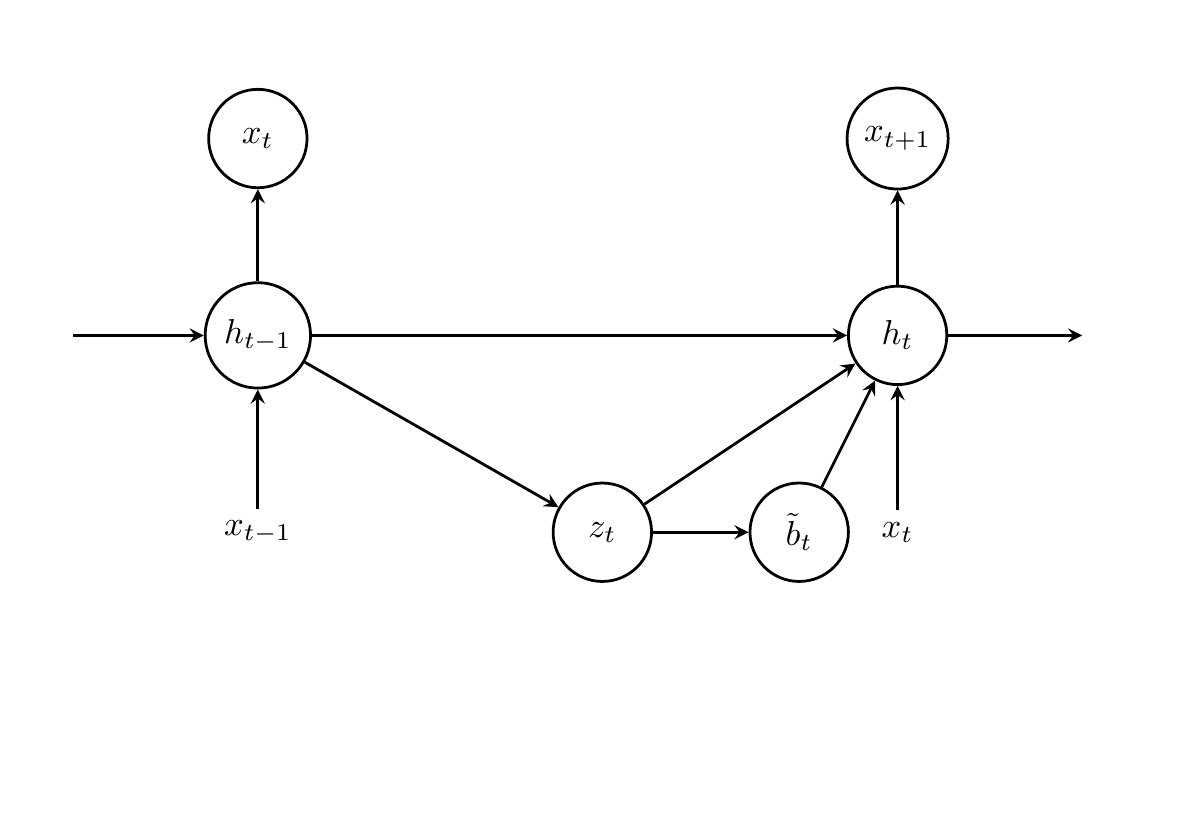}
  }
  \caption[]{ Graphical description of our proposed variational Bi-LSTM model during train phase (left) and inference phase (right). During training, each step $t$ is composed of an encoder which receives both the past and future summary via  $\*h_{t-1}$ and $\*b_t$ respectively, and a decoder that generates $\tilde{\*h}_{t-1}$ and $\tilde{\*b}_t$ which are forced to be close enough  to ${\*h}_{t-1}$ and ${\*b}_t$ using two auxiliary reconstruction costs (dashed lines). This dependence between backward and forward LSTM through the latent random variable encourages the forward LSTM to learn a richer representation. During inference, the backward LSTM is removed. In this case, $\*z_{t}$ is sampled from the prior as in a typical VAE, which in our case, is defined as a function of $\*h_{t-1}$. 
}
  \label{fig:model}
\end{figure}

In the forward LSTM model, we introduce additional latent random variables, $\*z_t$ and $\tilde{\*b}_t$, where $\*z_t$ depends on $\*h_{t-1}$ and $\*b_{t}$ during training, and $\tilde{\*b}_t$ depends on $\*z_t$ (see figure \ref{fig:model}-left, for a graphical representation). We also introduce the random variable $\tilde{\*h}_{t-1}$ which depends on $\*z_t$ and is used in an auxiliary cost which we will discuss later. Note that so far, $\tilde{\*b}_t$ and $\tilde{\*h}_{t-1}$ are simply latent vectors drawn from conditional distributions that depend on $\*z_t$, to be defined below.
%$p_{\psi}$ and $p_{\xi}$, respectively and not ${\*b}_t$ reconstructions. 
However, as explained in Section \ref{subsec:inf} (see also dashed lines in figure \ref{fig:model}-left), we will encourage these to lie near the manifolds of backward and forward LSTM states respectively by adding auxiliary costs to our objective.
% One other thing: I don't see \tilde{h} defined anywhere in the text, but it appears in the diagram. It doesn't appear in any equations

% $\*z=(\*z_1, \ldots, \*z_T)$ and  $\tilde{\*b} = (\tilde{\*b}_1, \ldots, \tilde{\*b}_{T})$ be two latent variable
% vectors.  
By design, the joint conditional distribution $p_{\theta, \psi}(\*z_{t}, \tilde{\*b}_t | \*x_{1:t}) $ over latent variables $\*z_t$ and $\tilde{\*b}_t$ with parameters $\theta$ and $\psi$ factorizes as  
$p_{\theta}(\*z_{t} | \*x_{1:t})  p_{\psi}(\tilde{\*b}_t|\*z_t)$.
%  $p_{\theta}(\*z_{t} | \*x_{1:t-1},\*z_{1:t-1})  p_{\psi}(\tilde{\*b}_t|\*z_t)$.
This factorization enables us to formulate several helpful auxiliary costs, as defined in the next subsection.
Further, $p_\eta(\*x_{t+1} | \*x_{1:t}, \*z_{t}, \tilde{\*b}_t)$ defines the generating model, which induces the distribution over the next observation given the previous states and the current input.

%in which   $\tilde{\*b} = (\tilde{\*b}_1, \ldots, \tilde{\*b}_{T})$ is sampled from a  conditional distribution $p_\psi$ and  $\*z=(\*z_1, \ldots, \*z_T)$ is a sequence of stochastic latent variables drawn 
%from either a conditional inference model $ q_\phi(\*z_{t} | \*x)$ or a conditional prior $p_\theta$. 
Then the marginal likelihood of each individual sequential data sample $\*x$ can be written as 
\begin{equation}
\begin{split}
    p(\*x; \boldsymbol{\Gamma}) &= \prod_{t=0}^T p(\*x_{t+1}|\*x_{1:t}) \\ %=\prod_{t=0}^T\int_{\*z_{1:T}}p(\*x_{t+1}|\*x_{1:t}, \*z_{1:t})p_{\theta}(\*z_{t} | \*x_{1:t-1}, \*z_{1:t-1})  d\*z_{1:T}\cr     %&=\int_{z_{1:T}}\prod_{t=0}^Tp(\*x_{t+1}|\*x_{1:t}, \*z_{1:t})p_{\theta}(\*z_{t} | \*x_{1:t-1}, \*z_{1:t-1}) dz_{1:T}\cr
    &=\prod_{t=0}^T\int_{\*z_{t}}\int_{\tilde{\*b}_t} \Big[p_{\eta}(\*x_{t+1}| \*x_{1:t}, \*z_{t}, \tilde{\*b}_t) p_{\psi}(\tilde{\*b}_t|\*z_{t})  p_{\theta}(\*z_{t} | \*x_{1:t})\Big]d\tilde{\*b}_t  d\*z_{t},\\
\end{split}
\label{eq:pjoint}
\end{equation} 
% \begin{equation}
% \begin{split}
%     p(\*x; \boldsymbol{\Gamma}) &= \prod_{t=0}^Tp(\*x_{t+1}|\*x_{1:t}) \\ %=\prod_{t=0}^T\int_{\*z_{1:T}}p(\*x_{t+1}|\*x_{1:t}, \*z_{1:t})p_{\theta}(\*z_{t} | \*x_{1:t-1}, \*z_{1:t-1})  d\*z_{1:T}\cr     %&=\int_{z_{1:T}}\prod_{t=0}^Tp(\*x_{t+1}|\*x_{1:t}, \*z_{1:t})p_{\theta}(\*z_{t} | \*x_{1:t-1}, \*z_{1:t-1}) dz_{1:T}\cr
%     &=\prod_{t=0}^T\int_{\*z_{1:T}}\int_{\tilde{\*b}_t} \Big[p_{\eta}(\*x_{t+1}| \*x_{1:t}, \*z_{1:t}, \tilde{\*b}_t) p_{\psi}(\tilde{\*b}_t|\*z_{1:t})  p_{\theta}(\*z_{t} | \*x_{1:t-1}, \*z_{1:t-1})\Big]d\tilde{\*b}_t  d\*z_{1:T},\\
% \end{split}
% \label{eq:pjoint}
% \end{equation} 
where $\boldsymbol{\Gamma} =\{\boldsymbol{\phi}, \boldsymbol{\theta}, \boldsymbol{\psi},  \boldsymbol{\eta}\}$ is the set of all parameters of the model. Here, we assume that all conditional distributions belong to parametrized families of distributions which can be evaluated and sampled from efficiently.

Note that the joint distribution in equation \eqref{eq:pjoint} is intractable. \citet{vae} demonstrated how to maximize a variational lower bound of the likelihood function. Here we derive a similar lower bound for the joint likelihood $\log p(\*x; \boldsymbol{\Gamma})$  given as ${\mathcal{L}}_{\boldsymbol{\Gamma}}$, of the data log likelihood, which is given by
\begin{align}
   \log p(\*x; \boldsymbol{\Gamma}) 
    \geq{\mathcal{L}}_{\boldsymbol{\Gamma}}=\sum_{t=0}^T\underset{{\*z_{t}\sim  q_\phi(\*z_{t} | \*x_{1:t})}}{\mathbb{E}}&\underset{{\tilde{\*b}_t\sim p_{\psi}(\tilde{\*b}_t|\*z_t)}}{\mathbb{E}}\Big[\log p_{\eta}(\*x_{t+1}| \*x_{1:t}, \*z_{t}, \tilde{\*b}_t) \Big] \\ 
    &\quad - D_{KL}(q_\phi(\*z_{t}|\*x_{1:t})\| p_{\theta}(\*z_{t} | \*x_{1:t})),
\label{eq:elbow}
\end{align}

%  \begin{equation}
% \begin{split}
%   \log p(\*x; \boldsymbol{\Gamma}) 
%     \geq{\mathcal{L}}_{\boldsymbol{\Gamma}}=\sum_{t=0}^T\underset{{\*z_{1:T}\sim  q_\phi(\*z | \*x)}}{\mathbb{E}}&\underset{{\tilde{\*b}_t\sim p_{\psi}(\tilde{\*b}_t|\*z_t)}}{\mathbb{E}}\Big[\log p_{\eta}(\*x_{t+1}| \*x_{1:t}, \*z_{1:t}, \tilde{\*b}_t) \Big] \\ 
%     &\quad - D_{KL}(q_\phi(\*z_{t}|\*x)\| p_{\theta}(\*z_{t} | \*x_{1:t}, \*z_{1:t-1})),
% \end{split}
% \label{eq:elbow}
% \end{equation}
%  \begin{equation}
% \begin{split}
%   \log p(\*x; \boldsymbol{\Gamma}) 
%     \geq{\mathcal{L}}_{\boldsymbol{\Gamma}}=\sum_{t=0}^T\underset{{\*z_{1:T}\sim  q_\phi(\*z | \*x)}}{\mathbb{E}}&\underset{{\tilde{\*b}_t\sim p_{\psi}(\tilde{\*b}_t|\*z_t)}}{\mathbb{E}}\Big[\log p_{\eta}(\*x_{t+1}| \*x_{1:t}, \*z_{1:t}, \tilde{\*b}_t) \Big] \\ 
%     &\quad - D_{KL}(q_\phi(\*z_{t}|\*x)\| p_{\theta}(\*z_{t} | \*x_{1:t-1}, \*z_{1:t-1})),
% \end{split}
% \label{eq:elbow}
% \end{equation}
 where $ q_\phi(\*z_{t} | \*x)$ is the  conditional inference model, $D_{KL}$  is the Kullback-Leibler (KL) divergence between the approximate posterior and the conditional prior (see the appendix). Notice the above function ${\mathcal{L}}_{\boldsymbol{\Gamma}}$ is a general lower bound that is not explicitly defined in terms of $\*h_t$ and $\*b_t$, but rather all the terms are conditional upon the previous predictions $\*x_{1:t}$. The choice of how the model is defined in terms of $\*h_t$ and $\*b_t$ is a design choice which we will make more explicit in the next section.

\subsection{Training and Inference}\label{subsec:inf}

In the proposed variational Bi-LSTM, the latent variable $\*z_t$ is inferred as
\begin{align}
\*z_t \sim q_\phi(\*z_{t} | (\*h_{t-1}, \*b_{t})) = \mathcal{N}(\boldsymbol{\mu}_{q,t}, \textrm{diag}(\boldsymbol{\sigma}_{q,t}^{2})),
\end{align}
in which  $[\boldsymbol{\mu}_{q,t}, \boldsymbol{\sigma}^2_{q,t}] = f_{\phi}(\*h_{t-1}, \*b_{t})$ where $f_{\phi}$ is a multi-layered feed-forward network with Gaussian outputs.
% to be either a standard multivariate  Gaussian distribution $\mathcal{N}(\*0, \*I)$ or  a
We assume that the prior over $\*z_{t}$ is a diagonal multivariate Gaussian distribution given by
\begin{equation}
  p_\theta(\*z_t | \*h_{t-1}) = \mathcal{N}(\boldsymbol{\mu}_{p,t}, \textrm{diag}(\boldsymbol{\sigma}_{p,t} ^{2})), \quad \textrm{where}  \quad [\boldsymbol{\mu}_{p,t}, \boldsymbol{\sigma}^2_{p,t}] = f_{\theta}(\*h_{t-1}),
\end{equation}
% \begin{equation}
%   p_\theta(\*z_t | \*x_{1:t}, \*z_{1:t-1}) = \mathcal{N}(\boldsymbol{\mu}_{p,t}, \textrm{diag}(\boldsymbol{\sigma}_{p,t} ^{2})), \quad \textrm{where}  \quad [\boldsymbol{\mu}_{p,t}, \boldsymbol{\sigma}^2_{p,t}] = f_{\theta}(\*h_{t-1}),
% \end{equation}
for a fully connected network  $f_{\theta}$. This is important because, during generation (see Figure \ref{fig:model}-right, for a graphical representation), we will not have access to the backward LSTM. In this case, as in a VAE, we will sample $\*z_t$ from the prior during inference which only depends on the forward path. Since we define the prior to be a function of $\*h_{t-1}$, the forward LSTM is encouraged during training to learn the dependence due to the backward hidden state $\*b_t$.

The latent variable $\tilde{\*b}_{t}$ is meant to model information coming from the future of the sequence.
Its conditional distribution is given by 
\begin{equation}
    p_{\psi}(\tilde{\*b}_{t} | \*z_t) = \mathcal{N}(\boldsymbol{\mu}_{\tilde{\*b},t}, \textrm{diag}(\boldsymbol{\sigma}_{\tilde{\*b},t}^{2})),
\end{equation}
where $[\boldsymbol{\mu}_{\tilde{\*b},t}, \boldsymbol{\sigma}^2_{\tilde{\*b},t}] = f_{\psi}(\*z_{t})$ for a fully connected neural network $f_{\psi}$ (See Figure \ref{fig:tr}). 
To encourage the encoding of future information in $\tilde{\*b}_{t}$, we maximize the probability of the true backward hidden state, $\*b_t$, under the distribution $p_\psi$,
as an auxiliary cost during training. In this way we treat $\tilde{\*b}_{t}$ as a predictor of $\*b_t$, similarly to what was done by~\citet{Sordoni2017}.

To capture information from the past in the latents, we similarly use $\tilde{\*h}_{t-1}$ as a predictor of $\*h_{t-1}$. This is accomplished by maximizing the probability of the latter under the conditional distribution of the former, $\log p_\xi (\tilde{\*h}_{t-1} | \*z_{t})$, as another auxiliary cost, where
\begin{equation}
    p_{\xi}(\tilde{\*h}_{t-1} | \*z_t) = \mathcal{N}(\boldsymbol{\mu}_{\tilde{\*h},t}, \textrm{diag}(\boldsymbol{\sigma}_{\tilde{\*h},t}^{2})).
\end{equation}
%In fact,to guarantee $\tilde{\*b}_{t}$ is close to backward path and the second one is $\log p_\xi (\*h_{t-1} | \*z_{t})$ where 
Here,  $[\boldsymbol{\mu}_{\tilde{\*h},t}, \boldsymbol{\sigma}^2_{\tilde{\*h},t}]$ is the output of a fully-connected neural network $f_{\xi}$ taking $\*z_{t}$ as input.
% and for simplicity  we assume $p_\xi(\*h | \*z) = \prod_{t=1}^T p_\xi(\*h_t | \*z_t)$ .
The auxiliary costs arising from distributions $p_\xi$ and $p_\psi$ teach the variational Bi-LSTM to encode past and future information into the latent space of $\*z$.

We assume that the generating distribution $p_\eta(\*x_{t+1} | \*x_{1:t}, \*z_{1:t}, \tilde{\*b}_t)$ is parameterized via a recurrent fully connected network, taking the form of either %${\boldsymbol{\mu}, \boldsymbol{\sigma}^2}$ parameters of
%with 
a Gaussian distribution output in the continuous case or categorical proportions output in the discrete (ie, one-hot) prediction case.% (see Figure~\ref{fig:inf}).

% All the parameters in  $\boldsymbol{\Gamma}$ and $\boldsymbol{\xi}$ are updated  based on backpropagation through time~\citep{rumelhart1988learning} using the reparameterization trick~\citep{vae}, where the gradients are computed by differentiating the following function:
Finally, we define the Variational Bi-LSTM objective we use in this paper by instantiating the conditionals upon $\*x_{1:t}$ used in Eq. \ref{eq:elbow} with functions $\*h_t$ and $\*b_t$ as,
% \begin{equation*}%\label{eq:loss}
% \begin{split}
%     \mathcal{L}(\*x; \boldsymbol{\Gamma},\boldsymbol{\xi}) =\sum_{t=0}^T\underset{{z_{1:T}\sim  q_\phi(\*z | \*x)}}{\mathbb{E}}\Big[&\ \underset{{\tilde{\*b}_t\sim  p_\psi(b|\*z_{1:t})}}{\mathbb{E}}\Big[\log p(\*x_{t+1}| \*x_{1:t}, \*z_{1:t}, \tilde{\*b}_t)+\\ 
%      &    \alpha \log p_\psi (\*b_{t} | \*z_{t})+\beta \log p_\xi (\*h_{t-1} | \*z_{t})\Big]\Big]- \\ 
%     & D_{KL}(q_\phi(\*z_{t}|\*x)\| p_{\theta}(\*z_{t} | \*x_{1:t}, \*z_{1:t-1})).\\
% \end{split}
% \label{eq:elbo}
% \end{equation*}
\begin{align}%\label{eq:loss}
\begin{split}
    \mathcal{L}(\*x; \boldsymbol{\Gamma},\boldsymbol{\xi}) =\sum_{t=0}^T\underset{{\*z_{t}\sim  q_\phi(\*z_t | \*h_{t-1}, \*b_t)}}{\mathbb{E}}\Big[&\ \underset{{\tilde{\*b}_t\sim  p_\psi(\tilde{\*b}_t|\*z_{t})}}{\mathbb{E}}\Big[\log p_\eta(\*x_{t+1}| \*h_{t}) + \log p(\*x_{t+1}| {\*b}_t) +\\ 
     &    \alpha \log p_\psi (\*b_{t} | \*z_{t})+\beta \log p_\xi (\*h_{t-1} | \*z_{t})\Big]\Big]- \\ 
    & D_{KL}(q_\phi(\*z_{t}|\*h_{t-1}, \*b_t)\| p_{\theta}(\*z_{t} | \*h_{t-1})),\\
\end{split}
\label{eq:elbo}
\end{align}
where $\alpha$ and $\beta$ are non-negative real numbers denoting the coefficients of the auxiliary costs $\log p_\psi (\*b_{t} | \*z_{t})$ and $\log p_\xi (\*h_{t-1} | \*z_{t})$ respectively. These auxiliary costs ensure that $\tilde{\*h}_t$ and $\tilde{\*b}_t$ remain close to $\*h_t$ and $\*b_t$. All the parameters in  $\boldsymbol{\Gamma}$ and $\boldsymbol{\xi}$ are updated  based on backpropagation through time~\citep{rumelhart1988learning} using the reparameterization trick~\citep{vae}.

As a side note, we improve training convergence with a trick which we refer to as stochastic backprop, meant to ease learning of the latent variables.
It is well known that autoregressive decoder models tend to ignore their stochastic variables~\citep{bowman2015generating}.
%which in general can be suitable for models that combine LSTMs with VAEs. 
Stochastic backprop is a technique to encourage that relevant summaries of the past and the future are encoded in the latent space. The idea is to stochastically skip gradients of the auxiliary costs \emph{with respect to the recurrent units} from backpropagating through time. To achieve this, at each time step, a mask drawn from a Bernoulli distribution which governs whether to skip the gradient or to backpropagate it for a given data point.

\section{Experimental results}\label{sec:exp}
In this section we demonstrate the effectiveness of our proposed model on several tasks.
We present experimental results obtained
when training Variational Bi-LSTM on various sequential datasets: Penn Treebank (PTB),  IMDB, TIMIT, Blizzard, and Sequential MNIST. 
Our main goal is to ensure that the model proposed in Section \ref{sec:md} can  benefit from a generated  relevant summary of the future that yields   competitive results.
In all experiments, we train all the models using ADAM optimizer~\citep{Kingma2014}  and we set all MLPs in Section \ref{sec:md}  to have one hidden layer with leaky-ReLU hidden activation.
%and with hidden units we refer to the number of  hidden layers units.  
All the models are implemented using Theano~\citep{2016arXiv160502688short} and the code is available at \url{https://anonymous.url}.
%\url{https://github.com/shabanian/????} \todo{clean code}.

  {\bf{Blizzard:}}  %To ascertain that our model works in general, we applied it to two speech modeling datasets. First one is 
 Blizzard is a speech model dataset with 300 hours of English, spoken by a single female speaker. We report  the average log-likelihood  for half-second sequences~\citep{srnn}. In our experimental setting, we use 1024 hidden units for MLPs, 1024 LSTM units and  512 latents. Our model is trained using learning rate of   0.001 and minibatches of size 32 and we set $\alpha=\beta=1$. A fully factorized multivariate Gaussian distribution  is used as the output distribution. The final lower bound estimation on TIMIT can be found in Table~\ref{tab:blitim}. 
 \begin{table}[h]
\caption{ The average of log-likelihood per sequence on Blizzard and TIMIT testset}
\label{tab:blitim}
\begin{center}
\begin{tabular}{lcc}
\multicolumn{1}{c} { \textbf{Model} } &\multicolumn{1}{c}{\textbf{Blizzard}} &\multicolumn{1}{c}{\textbf{TIMIT}}
\\ \hline \\ 
        %   \midrule
          RNN-Gauss & 3539 & -1900 \\
          RNN-GMM & 7413 & 26643 \\
          VRNN-I-Gauss & $\ge$ 8933 & $\ge$ 28340 \\ 
          VRNN-Gauss & $\ge$ 9223 & $\ge$ 28805 \\
          VRNN-GMM & $\ge$ 9392 & $\ge$ 28982 \\
          SRNN (smooth+res$_q$) & $\ge$ 11991 & $\ge$ 60550 \\
          Z-Forcing {\small~\citep{Sordoni2017}}  & $\ge$ 14315 & $\ge$ 68852 \\
                    % \midrule
          Variational Bi-LSTM    & $\ge$   \textbf{17319} & $\ge$    \textbf{73976}\\
\end{tabular}
\end{center}
\end{table}
 
%The trick  which we refer to as skipping gradient, is  basically not back-propagate the gradient with  disconnecting stochastically the target of VAE module, i.e. $\*b$ and  $\*h$ during training  using a mask drawn from Bernoulli distribution of mini-batch size. Rescaling  is a great help to achieve the reported results in Section~\ref{sec:exp} and it is meant to allow both deterministic and stochastic parts of the model learn efficiently and specially not to ignore the stochastic part due to noise. 

  {\bf{TIMIT:}} Another speech modeling dataset is  TIMIT  with 6300 English sentences read by 630 speakers. Like ~\citet{srnn}, our model is trained on raw sequences of 200 dimensional frames. % We use the same procedure for validation and the average log-likelihood for the sequences is reported in the test-set. 
  In our experiments, we  use  1024 hidden units, 1024 LSTM units and  128 latent variables, and batch size of 128.  We train the model using learning rate of 0.0001,  $\alpha=0.001$ and $\beta=0$. The average log-likelihood for the sequences on test can be found in Table~\ref{tab:blitim}.

{\bf{Sequential MNIST:}} We use the MNIST dataset which is binarized according to~\citet{murray2009latent} and we download it from~\citet{mnistla}. Our best model consists of  1024 hidden units, 1024 LSTM units and 256 latent variables. We train the model using a learning rate of 0.0001 and a batch size of 32. To reach the negative log-likelihood reported in Table~\ref{tab:seqmnist}, we set $\alpha= 0.001$ and $\beta=0$.  

\begin{table}[h]
\caption{The average of negative log-likelihood on sequential MNIST}
\label{tab:seqmnist}
\begin{center}
\begin{tabular}{lc}
\multicolumn{1}{c} { \textbf{Models} } &\multicolumn{1}{c}{\textbf{Seq-MNIST}}  
\\ \hline \\ 
        %   \midrule
          DBN 2hl{\small ~\citep{germain2015made}} &     $\approx$ 84.55 \\
          NADE{\small~\citep{uria2016neural}}  &      88.33 \\
          EoNADE-5 2hl{\small~\citep{raiko-nips2014}} &     84.68 \\
          DLGM 8 ~\citep{salimans2014markov}       &       $\approx$ 85.51 \\
          DARN 1hl~\citep{gregor2015draw} &     $\approx$ 84.13 \\
           BiHM~\citep{DBLP:journals/corr/BornscheinSFB15} &     $\approx$ 84.23 \\
          DRAW~\citep{gregor2015draw}        &      $\leq$ 80.97 \\
          PixelVAE{\small~\citep{gulrajani2016pixelvae}} & $\approx$ \textbf{79.02}$^\blacktriangledown$  \\
          Prof. Forcing{\small~\citep{DBLP:conf/nips/GoyalLZZCB16}}  & 79.58$^\blacktriangledown$ \\
          PixelRNN$_{\text{(1-layer)}}${\small~\citep{oord2016pixel}}           & 80.75 \\
          PixelRNN$_{\text{(7-layer)}}${\small~\citep{oord2016pixel}}           & 79.20$^\blacktriangledown$ \\
          Z-Forcing{\small~\citep{Sordoni2017}} & $\le$ 80.09  \\
                    %  \midrule
Variational Bi-LSTM       & $\le$ 79.78 \\
\end{tabular}
\end{center}
\end{table}

{\bf{IMDB:}} It is a dataset consists of 350000 movie reviews~\citep{diao2014jointly} in which each sentence has less than 16 words and the vocabulary size is fixed to 16000 words. % Following the setting described in~\cite{hu2017controllable}, 
In this experiment, we use 500  hidden units, 500 LSTM units and latent variables of size 64. The model is trained with a batch size of 32 and a learning rate of 0.001 and we set $\alpha=\beta=1$. The word perplexity  on valid and test dataset is shown in Table~\ref{tab:imdb}.  
\begin{table}[h]  
\caption{Word perplexity on IMDB on valid and test sets}
\label{tab:imdb}
\begin{center}
\begin{tabular}{lcc}
\multicolumn{1}{c}{\bf Model}  &\multicolumn{1}{c}{\bf Valid} &\multicolumn{1}{c}{\bf Test}
\\ \hline \\
Gated Word-Char & 70.60& 70.87\\
Z-Forcing{\small~\citep{Sordoni2017}} & 56.48& 65.68\\
Variational Bi-LSTM    &   \textbf{51.43} & \textbf{51.60} \\
\end{tabular}
\end{center}
\end{table}

{\bf{PTB:}} Penn Treebank (\cite{Marcus:1993:BLA:972470.972475}) is a language model dataset consists of 1 million words. We train our model with 1024 LSTM units, 1024 hidden units, and the latent variables of size  128.  We train the model using a  standard Gaussian prior, a learning rate of 0.001 and batch size of 50 and we set $\alpha=\beta=1$. The model is trained to predict the next character in a
sequence and the final bits per character on test and valid sets are shown in Table~\ref{tab:chptb}.

%For some set of datasets, such as PTB  choosing approximate inference equal to prior is sufficient to yield a good performance.
%In this setting, our models use a fully factorized multivariate Gaussian distribution as the output distribution for each time-step. 

\begin{table}[h]
\caption{Bits Per Character (BPC) on PTB valid and test sets}
\label{tab:chptb}
\begin{center}
\begin{tabular}{lcc}
\multicolumn{1}{c}{\bf Model}  &\multicolumn{1}{c}{\bf Valid} &\multicolumn{1}{c}{\bf Test}
\\ \hline \\
Unregularized LSTM         & 1.47& 1.36 \\
Weight noise & 1.51& 1.34 \\
Norm stabilizer & 1.46 & 1.35\\
Stochastic depth & 1.43 &  1.34 \\
Recurrent dropout& 1.40& 1.29\\
Zoneout (\cite{DBLP:journals/corr/KruegerMKPBKGBL16})& 1.36& 1.25\\
RBN (\cite{DBLP:journals/corr/CooijmansBLC16}) & -  & 1.32\\
H-LSTM + LN (\cite{DBLP:journals/corr/HaDL16}) & 1.28& 1.25\\
3-HM-LSTM + LN (Chung et al., 2016)         & - & 1.24\\
%HWN (\cite{DBLP:journals/corr/ZillySKS16}) &-&1.26\\
2-H-LSTM + LN (\cite{DBLP:journals/corr/HaDL16}) & 1.25 & \textbf{1.22}\\ 
Z-Forcing    & 1.29 & 1.26\\
Variational Bi-LSTM          & 1.26 & 1.23 \\
\end{tabular}
\end{center}
\end{table}

\iffalse
\begin{figure}[t]
\centering
\subfigure{
 \includegraphics[scale=0.4]{imdb.png}
}
\subfigure{
 \includegraphics[scale=0.4]{ptb.png}
 }
 \caption[]{ 
Valid and test perplexity on IMDB and bits per character on PTB
}
  \label{fig:imdb}
\end{figure}
\fi
%-------------------------------

%\begin{table}[h]
%\caption{Final performance with and without using encoder \samira{not sure if  all models are saved TODO }}
%\label{tab:enccom}
%\begin{center}
%\begin{tabular}{cccccc}
%\multicolumn{1}{c} { \textbf{Dataset} } %&\multicolumn{1}{c}{\textbf{PTB}}&\multicolumn{1}{c}{\textbf{Seq-MNIST}}   %&\multicolumn{1}{c}{\textbf{IMDB}} &\multicolumn{1}{c}{\textbf{TIMIT}} %&\multicolumn{1}{c}{\textbf{Blizzard}}  
%\\ \hline \\ 
%          \midrule
%        \textbf{With encoder }&1.23&     & 51& &  \\
%           \textbf{Without encoder} &1.23  &     &52 & &  \\
%\end{tabular}
%\end{center}
%\end{table}

\section{Ablation Studies}
 \label{sec_ablation}
 The goal of this section is to study the importance of the various components in our model and ensure that these components provide performance gains. The experiments are as follows:

 \begin{table}[h]
\caption{Perplexity on IMDB using different coefficient $\gamma$ for activity regularization}
\label{tab:pplgam}
\begin{center}

\begin{tabular}{cccccc}
\multicolumn{1}{c} { \textbf{$\gamma$ } } &\multicolumn{1}{c}{\textbf{0.001}} &\multicolumn{1}{c}{\textbf{1.}}&\multicolumn{1}{c}{\textbf{4.}}   &\multicolumn{1}{c}{\textbf{8.}} &\multicolumn{1}{c}{\textbf{16.}}  
\\ \hline \\ 
        %   \midrule
        \textbf{Test perplexity}&56.07&60.74&  69.97  &77.24 &86.72 \\
        \label{table_ar_ablation}
\end{tabular}
\end{center}
\end{table}

\textbf{1. Reconstruction loss on $\*h_t$ vs activity regularization on $\*h_t$}

\cite{ar_tar} study the importance of activity regularization (AR) on the hidden states of LSTMs given as,
\begin{align}
\mathcal{R}_{AR} &= {\gamma} \lVert \mathbf{h}_t \rVert_2^2.
\end{align} 
Since our model's reconstruction term on $\mathbf{h}_t$ can be decomposed as,
\begin{align}
\lVert \*h_t - \tilde{\*h}_t \rVert_2^2 &= \lVert \*h_t \rVert^2_2  +\lVert \tilde{\*h}_t \rVert^2_2 - 2  \*h_t^T \tilde{\*h}_t 
\end{align}
we perform experiments to confirm that the gains in our approach is not due to the $\ell^2$ regularization alone since our regularization encapsulates an $\ell^2$ term along with the dot product term.

To do so, we replace the auxiliary reconstruction terms in our objective with activity regularization using hyperparameter $\alpha \in \{0.001, 1, 4, 8, 16 \}$ and study the test perplexity. The results are shown in table \ref{table_ar_ablation}. We find that in all the cases performance using activity regularization is worse compared with our best model shown in table \ref{tab:imdb}.

 \begin{table}[h]
\caption{KL divergence of the Variational Bi-LSTM}
\label{tab:kld}
\begin{center}
\begin{tabular}{cccccc}
\multicolumn{1}{c} { \textbf{Dataset} } &\multicolumn{1}{c}{\textbf{PTB}}&\multicolumn{1}{c}{\textbf{Seq-MNIST}}   &\multicolumn{1}{c}{\textbf{IMDB}} &\multicolumn{1}{c}{\textbf{TIMIT}} &\multicolumn{1}{c}{\textbf{Blizzard}}  
\\ \hline \\ 
          %  \midrule
        \textbf{KL}&0.001 &0.02     &0.18 &3204.71 &3799.79  \\
\end{tabular}
\end{center}
\end{table}

\begin{figure}[t]
  \centering
 % \subfigure{
 %   \includegraphics[scale=0.3]{bli_valid.png}
 % }
    \subfigure{
    \includegraphics[scale=0.4]{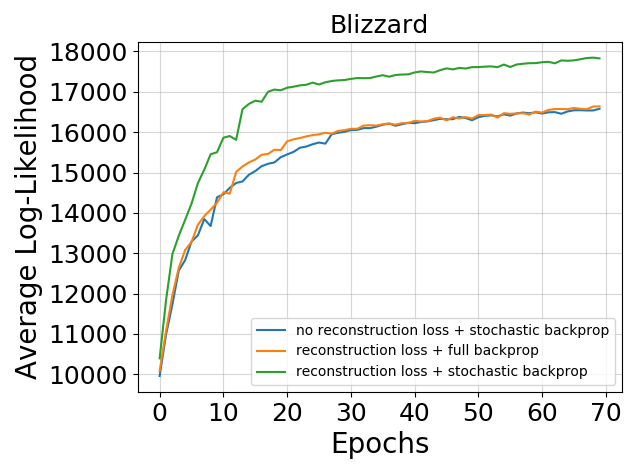}
  }
 % \caption[]{Test average log-likelihood on Blizzard 
%}
 % \subfigure{
 %   \includegraphics[scale=0.3]{bli_valid.png}
 % }
    \subfigure{
    \includegraphics[scale=0.4]{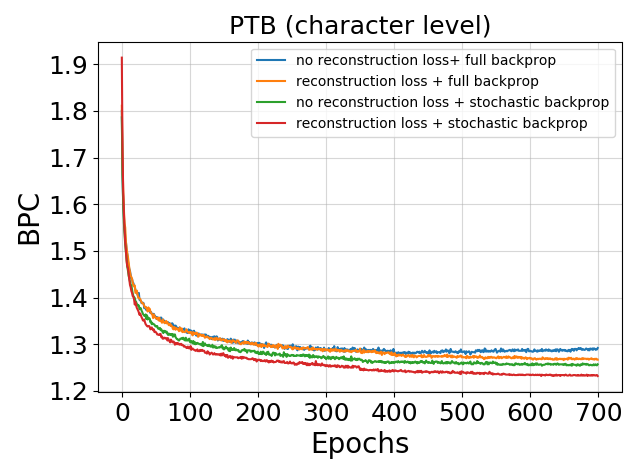}
  }
  \caption[]{Evolution of validation set performance during training of Variational Bi-LSTMs with and without auxiliary reconstruction costs and stochastic backprop through auxiliary costs on PTB and Blizzard. We see that both the presence of reconstruction loss and stochastic back-propagation through them helps performance.
 }
 \label{fig:bli}
\end{figure}

\begin{figure}[t]
  \centering
 % \subfigure{
 %   \includegraphics[scale=0.3]{bli_valid.png}
 % }
    \subfigure{
    \includegraphics[scale=0.4]{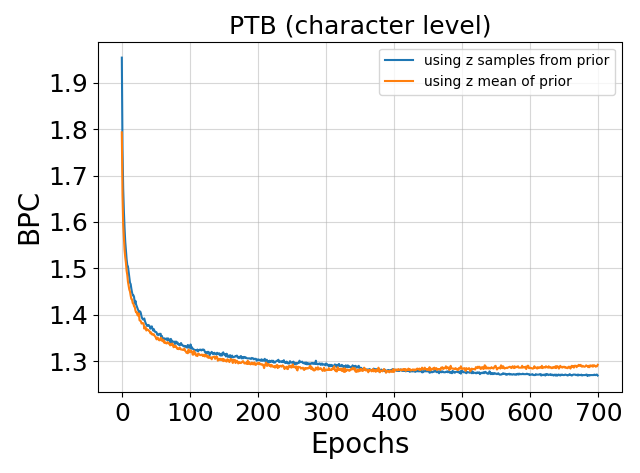}
  }
 % \caption[]{Test average log-likelihood on Blizzard 
%}
 % \subfigure{
 %   \includegraphics[scale=0.3]{bli_valid.png}
 % }
  \caption[]{Evolution of bits per character (BPC) on PTB validation set with sampling latent variables $\*z_t$ from  $q_{\phi}(\*z_t | \*h_{t-1}, \*b_t)$ during  training vs. using the expected value of $\*z_t$. Performance drops when $\*z_t$ is not sampled. This shows sampling $\*z_t$ randomly during training acts as a regularizer.}
  %{Evolution of bits per character (BPC) on PTB validation set with sampling latent variables $\*z$ from  $\mathcal{N}(\*0,\*I)$ during  training vs. using a fixed vector which we set to  be  the mean of latent variables. Interestingly, not sampling from prior during inference does not hurt the final performance on PTB.}
  %
  \label{fig:tpb-abl}
\end{figure}

\textbf{2. Use of parametric encoder prior vs. fixed Gaussian prior}

In our variational Bi-LSTM model, we propose to have the encoder prior over $\*z_t$ as a function of the previous forward LSTM hidden state $\*h_{t-1}$. This is done to omit the need of the backward LSTM during inference because it is unavailable in practical scenarios since predictions are made in the forward direction. However, to study whether the model learns to use this encoder or not, we record the KL divergence value of the best validation model for the various datasets. The results are reported in table \ref{tab:kld}. We can see that the KL divergence values are large in the case of IMDB, TIMIT and Blizzard datasets, but small in the case of Seq-MNIST and PTB. To further explore, we run experiments on these datasets with fixed standard Gaussian prior like in the case of traditional VAE. Interestingly we find that the model with fixed prior performed similarly in the case of PTB, but hurt performance in the other cases, which can be explained given their large KL divergence values in the original experiments.

\textbf{3. Effectiveness of auxiliary costs and stochastic back-propagation}

In our model description, we propose to stochastically back propagate gradients through the auxiliary (reconstruction) costs, i.e., randomly choose to pass the gradients of the auxiliary cost or not. Here we evaluate the importance of the auxiliary costs and stochastic back-propagation. Figure \ref{fig:bli} shows the evolution of validation performance on the Blizzard and PTB dataset. In both cases we see that both the auxiliary costs and stochastic back-propagation help the validation set performance.

\textbf{4. Importance of sampling from VAE prior during training}

We evaluate the effectiveness of sampling $\*z_t$ from the prior during training vs. using the mean of the Gaussian prior. The validation set performance during training is shown in figure \ref{fig:tpb-abl}. It can be seen that sampling $\*z_t$ leads to better generalization. Further, with the model trained with $\*z_t$ sampled, we also evaluate if using samples is necessary during inference or not. Interestingly we find that during inference, the performance is identical in both cases; thus the deterministic mean of the prior can be used during inference.

\section{Related Work}
\label{sec_related_work}
Variational auto-encoders~\citep{vae} can be easily combined with many deep learning models. They have been applied in the feed-forward setting but they have also found usage in RNNs to better capture variation in sequential data~\citep{Sordoni2017, srnn, vrnn, storn}. VAEs consists of several  muti-layer neural networks  as probabilistic encoders and  decoders and  training is based on the gradient on log-likelihood lower bound (as the likelihood is in general intractable) of the model parameters $\Gamma$ along with a reparametrization trick.  The derived variational lower-bound  $\mathcal{L}_{\Gamma}$ for an observed random variable  $\*x$ is:
\begin{align}
\log p(\*x) \ge \mathcal{L}_{\Gamma}=\underset{\*z \sim q_\phi(\*z | \*x)}{\mathbb{E}}\left[\log \frac{p(\*x, \*z)}{q_\phi(\*z|\*x)}\right] =\underset{\*z \sim  q_\phi(\*z | \*x)}{\mathbb{E}} \Big[\ln p_\theta(\*x | \*z)\Big] - D_{KL}(q_\phi(\*z|\*x)\|p_\theta(\*z)),
\label{eq:vaeelbo}
\end{align}
where $D_{KL}$ denotes the Kullback-Leibler divergence and $p_\theta$ is the prior over a latent variable $\*z$. %%The KL divergence term is a (asymmetric) measure of the difference between two distributions.
%It is always positive, and it is zero if and only if the distributions are the
%same. 
The KL divergence term can be expressed as the difference between the cross-entropy of the prior w.r.t. $q_\phi(\*z|\*x)$ and the entropy of $q_\phi(\*z|\*x)$, and fortunately, it can be analytically computed and differentiated for some distribution families like Gaussians. Although maximizing the log-likelihood corresponds to minimizing the KL divergence,  we have to ensure that the resulting $q_\phi$ remains far enough from an undesired equilibrium state where $q_\phi(\*z|\*x)$ is almost everywhere equal to the prior over latent variables. 
Combining  recurrent neural networks with variational auto encoders can lead to powerful generative models that are capable of capturing the variations in data, however, they suffer badly from this optimization issue as discussed by~\citet{bowman2015generating}.

However, VAEs have successfully been applied to Bi-LSTMs by~\citet{Sordoni2017} through a technique called Z-forcing. It is a powerful generative auto-regressive model which is trained using the following  variational evidence lower-bound 
\begin{equation*}
\begin{split}
   \mathcal{L}(\*x; \theta, \phi, \xi) = \sum_{t} \underset{q_\phi(\*z_{t} | \*x)}{\mathbb{E}} & \Big[\log p_\theta(\*x_{t + 1} | \*x_{1:t}, \*z_{1:t}) \Big] -  D_{KL}(q_\phi(\*z_{t}|\*x)\|p_{\theta}(\*z_{t} | \*x_{1:t-1}, \*z_{1:t-1}))
   \end{split}
    \end{equation*}
 plus an auxiliary cost as a regularizer which is defined as  $\log p_\xi (\*b_{t} | \*z_{t})$. It is shown that the auxiliary cost helps in improving the final performance; however during inference the backward reconstructions are not used in their approach. In our ablation study section below, we show experimentally that this connection is important for improving the performance of Bi-LSTMs as is the case in our model.
 
 Twin networks on the other hand is a deterministic method which enforces the hidden states of the forward and backward paths of a Bi-LSTM to be similar. Specifically, this is done by adding as a regularization $\ell^2$ norm of difference between the pair of hidden states at each time step. The intuition behind this regularization is to encourage the hidden representations to be compatible towards future predictions. Notice a difference between twin networks and our approach is that twin networks 
 forces the hidden states of both the LSTMs to be similar to each other while our approach directly feeds a latent variable $\tilde{\*b}_t$ to the forward LSTM that is trained to be similar to the backward LSTM hidden state ${\*b}_t$. Hence while twin networks discard information from the backward path during inference, our model encourages the use of this information. 
 %Our approach on the other hand uses approximate inference (of $\tilde{\*b}_t$), thereby indirectly using information from the backward path even though it is not actually used during inference.

% IF THERE IS TIME, CUT DOWN ON VAE DESCRIPTION (it's well known now) AND INSTEAD describe VRNN, SRNN, TwinNet 

\section{Conclusion} 
We propose Variational Bi-LSTM as an auto-regressive generative model by framing a joint objective that effectively creates a channel for exchanging information between the forward and backward LSTM. We achieve this by deriving a variational lower bound of the joint likelihood of the temporal data sequence. We empirically show that our variational Bi-LSTM model acts as a regularizer for the forward LSTM and leads to performance improvement on different benchmark sequence generation problems. We also show through ablation studies the importance of the various components in our model.

% Moreover, the conditional distribution over the backward LSTM variables is learned that can lead to better learning results in practice. Furthermore, Variational Bi-LSTM model acts as a regularizer and makes both networks be informative enough to perform well on different benchmark problems taken from the literature.

\section*{Acknowledgments}
The authors would like to thank Theano developers  ~\citep{2016arXiv160502688short} for their great work. DA was supported by IVADO, CIFAR and NSERC.
%3.  Adam 

\bibliography{iclr2018_conference}

\begin{thebibliography}{37}
\providecommand{\natexlab}[1]{#1}
\providecommand{\url}[1]{\texttt{#1}}
\expandafter\ifx\csname urlstyle\endcsname\relax
  \providecommand{\doi}[1]{doi: #1}\else
  \providecommand{\doi}{doi: \begingroup \urlstyle{rm}\Url}\fi

\bibitem[Arik et~al.(2017)Arik, Chrzanowski, Coates, Diamos, Gibiansky, Kang,
  Li, Miller, Raiman, Sengupta, et~al.]{arik2017deep}
Sercan~O Arik, Mike Chrzanowski, Adam Coates, Gregory Diamos, Andrew Gibiansky,
  Yongguo Kang, Xian Li, John Miller, Jonathan Raiman, Shubho Sengupta, et~al.
\newblock Deep voice: Real-time neural text-to-speech.
\newblock \emph{arXiv preprint arXiv:1702.07825}, 2017.

\bibitem[Bayer \& Osendorfer(2014)Bayer and Osendorfer]{storn}
Justin Bayer and Christian Osendorfer.
\newblock Learning stochastic recurrent networks.
\newblock \emph{arXiv preprint arXiv:1411.7610}, 2014.

\bibitem[Bornschein et~al.(2015)Bornschein, Shabanian, Fischer, and
  Bengio]{DBLP:journals/corr/BornscheinSFB15}
J{\"{o}}rg Bornschein, Samira Shabanian, Asja Fischer, and Yoshua Bengio.
\newblock Training opposing directed models using geometric mean matching.
\newblock \emph{CoRR}, abs/1506.03877, 2015.
\newblock URL \url{http://arxiv.org/abs/1506.03877}.

\bibitem[Bowman et~al.(2015)Bowman, Vilnis, Vinyals, Dai, Jozefowicz, and
  Bengio]{bowman2015generating}
Samuel~R Bowman, Luke Vilnis, Oriol Vinyals, Andrew~M Dai, Rafal Jozefowicz,
  and Samy Bengio.
\newblock Generating sentences from a continuous space.
\newblock \emph{arXiv preprint arXiv:1511.06349}, 2015.

\bibitem[Chung et~al.(2014)Chung, Gulcehre, Cho, and Bengio]{gru}
Junyoung Chung, Caglar Gulcehre, KyungHyun Cho, and Yoshua Bengio.
\newblock Empirical evaluation of gated recurrent neural networks on sequence
  modeling.
\newblock \emph{arXiv preprint arXiv:1412.3555}, 2014.

\bibitem[Chung et~al.(2015)Chung, Kastner, Dinh, Goel, Courville, and
  Bengio]{vrnn}
Junyoung Chung, Kyle Kastner, Laurent Dinh, Kratarth Goel, Aaron~C Courville,
  and Yoshua Bengio.
\newblock A recurrent latent variable model for sequential data.
\newblock In \emph{Advances in neural information processing systems}, pp.\
  2980--2988, 2015.

\bibitem[Cooijmans et~al.(2016)Cooijmans, Ballas, Laurent, and
  Courville]{DBLP:journals/corr/CooijmansBLC16}
Tim Cooijmans, Nicolas Ballas, C{\'{e}}sar Laurent, and Aaron~C. Courville.
\newblock Recurrent batch normalization.
\newblock \emph{CoRR}, abs/1603.09025, 2016.
\newblock URL \url{http://arxiv.org/abs/1603.09025}.

\bibitem[Diao et~al.(2014)Diao, Qiu, Wu, Smola, Jiang, and
  Wang]{diao2014jointly}
Qiming Diao, Minghui Qiu, Chao-Yuan Wu, Alexander~J Smola, Jing Jiang, and
  Chong Wang.
\newblock Jointly modeling aspects, ratings and sentiments for movie
  recommendation (jmars).
\newblock In \emph{Proceedings of the 20th ACM SIGKDD international conference
  on Knowledge discovery and data mining}, pp.\  193--202, 2014.

\bibitem[Fraccaro et~al.(2016)Fraccaro, S{\o}nderby, Paquet, and Winther]{srnn}
Marco Fraccaro, S{\o}ren~Kaae S{\o}nderby, Ulrich Paquet, and Ole Winther.
\newblock Sequential neural models with stochastic layers.
\newblock In \emph{Advances in Neural Information Processing Systems}, pp.\
  2199--2207, 2016.

\bibitem[Gal \& Ghahramani(2016)Gal and Ghahramani]{vdropout}
Yarin Gal and Zoubin Ghahramani.
\newblock A theoretically grounded application of dropout in recurrent neural
  networks.
\newblock In \emph{Advances in neural information processing systems}, pp.\
  1019--1027, 2016.

\bibitem[Germain et~al.(2015)Germain, Gregor, Murray, and
  Larochelle]{germain2015made}
Mathieu Germain, Karol Gregor, Iain Murray, and Hugo Larochelle.
\newblock Made: Masked autoencoder for distribution estimation.
\newblock In \emph{ICML}, pp.\  881--889, 2015.

\bibitem[Goyal et~al.(2016)Goyal, Lamb, Zhang, Zhang, Courville, and
  Bengio]{DBLP:conf/nips/GoyalLZZCB16}
Anirudh Goyal, Alex Lamb, Ying Zhang, Saizheng Zhang, Aaron~C. Courville, and
  Yoshua Bengio.
\newblock Professor forcing: {A} new algorithm for training recurrent networks.
\newblock In \emph{Advances in Neural Information Processing Systems 29: Annual
  Conference on Neural Information Processing Systems 2016, December 5-10,
  2016, Barcelona, Spain}, pp.\  4601--4609, 2016.
\newblock URL
  \url{http://papers.nips.cc/paper/6099-professor-forcing-a-new-algorithm-for-training-recurrent-networks}.

\bibitem[Gregor et~al.(2015)Gregor, Danihelka, Graves, Rezende, and
  Wierstra]{gregor2015draw}
Karol Gregor, Ivo Danihelka, Alex Graves, Danilo~Jimenez Rezende, and Daan
  Wierstra.
\newblock Draw: A recurrent neural network for image generation.
\newblock \emph{arXiv preprint arXiv:1502.04623}, 2015.

\bibitem[Gulrajani et~al.(2016)Gulrajani, Kumar, Ahmed, Taiga, Visin, Vazquez,
  and Courville]{gulrajani2016pixelvae}
Ishaan Gulrajani, Kundan Kumar, Faruk Ahmed, Adrien~Ali Taiga, Francesco Visin,
  David Vazquez, and Aaron Courville.
\newblock Pixelvae: A latent variable model for natural images.
\newblock \emph{arXiv preprint arXiv:1611.05013}, 2016.

\bibitem[Ha et~al.(2016)Ha, Dai, and Le]{DBLP:journals/corr/HaDL16}
David Ha, Andrew~M. Dai, and Quoc~V. Le.
\newblock Hypernetworks.
\newblock \emph{CoRR}, abs/1609.09106, 2016.
\newblock URL \url{http://arxiv.org/abs/1609.09106}.

\bibitem[Hochreiter \& Schmidhuber(1997)Hochreiter and Schmidhuber]{lstm}
Sepp Hochreiter and J{\"u}rgen Schmidhuber.
\newblock Long short-term memory.
\newblock \emph{Neural computation}, 9\penalty0 (8):\penalty0 1735--1780, 1997.

\bibitem[Kingma \& Ba(2014)Kingma and Ba]{Kingma2014}
Diederik Kingma and Jimmy Ba.
\newblock Adam: A method for stochastic optimization.
\newblock \emph{arXiv preprint arXiv:1412.6980}, 2014.

\bibitem[Kingma \& Welling(2014)Kingma and Welling]{vae}
Diederik~P Kingma and Max Welling.
\newblock {Stochastic Gradient VB and the Variational Auto-Encoder}.
\newblock \emph{2nd International Conference on Learning Representationsm
  (ICLR)}, pp.\  1--14, 2014.
\newblock ISSN 0004-6361.
\newblock \doi{10.1051/0004-6361/201527329}.
\newblock URL \url{http://arxiv.org/abs/1312.6114}.

\bibitem[Krueger et~al.(2016)Krueger, Maharaj, Kram{\'{a}}r, Pezeshki, Ballas,
  Ke, Goyal, Bengio, Larochelle, Courville, and
  Pal]{DBLP:journals/corr/KruegerMKPBKGBL16}
David Krueger, Tegan Maharaj, J{\'{a}}nos Kram{\'{a}}r, Mohammad Pezeshki,
  Nicolas Ballas, Nan~Rosemary Ke, Anirudh Goyal, Yoshua Bengio, Hugo
  Larochelle, Aaron~C. Courville, and Chris Pal.
\newblock Zoneout: Regularizing rnns by randomly preserving hidden activations.
\newblock \emph{CoRR}, abs/1606.01305, 2016.
\newblock URL \url{http://arxiv.org/abs/1606.01305}.

\bibitem[Larochelle(2011)]{mnistla}
Hugo Larochelle.
\newblock Binarized mnist dataset.
\newblock 2011.
\newblock URL \url{http://www.cs.toronto.edu/~larocheh/
  public/datasets/binarized\_mnist/ binarized_mnist_train.amat}.

\bibitem[Marcus et~al.(1993)Marcus, Marcinkiewicz, and
  Santorini]{Marcus:1993:BLA:972470.972475}
Mitchell~P. Marcus, Mary~Ann Marcinkiewicz, and Beatrice Santorini.
\newblock Building a large annotated corpus of english: The penn treebank.
\newblock \emph{Comput. Linguist.}, 19\penalty0 (2):\penalty0 313--330, June
  1993.
\newblock ISSN 0891-2017.
\newblock URL \url{http://dl.acm.org/citation.cfm?id=972470.972475}.

\bibitem[Mehri et~al.(2016)Mehri, Kumar, Gulrajani, Kumar, Jain, Sotelo,
  Courville, and Bengio]{mehri2016samplernn}
Soroush Mehri, Kundan Kumar, Ishaan Gulrajani, Rithesh Kumar, Shubham Jain,
  Jose Sotelo, Aaron Courville, and Yoshua Bengio.
\newblock Samplernn: An unconditional end-to-end neural audio generation model.
\newblock \emph{arXiv preprint arXiv:1612.07837}, 2016.

\bibitem[Merity et~al.(2017)Merity, McCann, and Socher]{ar_tar}
Stephen Merity, Bryan McCann, and Richard Socher.
\newblock Revisiting activation regularization for language rnns.
\newblock \emph{arXiv preprint arXiv:1708.01009}, 2017.

\bibitem[Murray \& Salakhutdinov(2009)Murray and
  Salakhutdinov]{murray2009latent}
Iain Murray and Ruslan~R Salakhutdinov.
\newblock Evaluating probabilities under high-dimensional latent variable
  models.
\newblock In D.~Koller, D.~Schuurmans, Y.~Bengio, and L.~Bottou (eds.),
  \emph{Advances in Neural Information Processing Systems 21}, pp.\
  1137--1144. Curran Associates, Inc., 2009.
\newblock URL
  \url{http://papers.nips.cc/paper/3584-evaluating-probabilities-under-high-dimensional-latent-variable-models.pdf}.

\bibitem[Oord et~al.(2016)Oord, Kalchbrenner, and Kavukcuoglu]{oord2016pixel}
Aaron van~den Oord, Nal Kalchbrenner, and Koray Kavukcuoglu.
\newblock Pixel recurrent neural networks.
\newblock \emph{arXiv preprint arXiv:1601.06759}, 2016.

\bibitem[Pascanu et~al.(2012)Pascanu, Mikolov, and
  Bengio]{DBLP:journals/corr/abs-1211-5063}
Razvan Pascanu, Tomas Mikolov, and Yoshua Bengio.
\newblock Understanding the exploding gradient problem.
\newblock \emph{CoRR}, abs/1211.5063, 2012.
\newblock URL \url{http://arxiv.org/abs/1211.5063}.

\bibitem[Raiko et~al.(2014)Raiko, Li, Cho, and Bengio]{raiko-nips2014}
Tapani Raiko, Yao Li, Kyunghyun Cho, and Yoshua Bengio.
\newblock Iterative neural autoregressive distribution estimator nade-k.
\newblock In \emph{Advances in neural information processing systems}, pp.\
  325--333, 2014.

\bibitem[Rumelhart et~al.(1988)Rumelhart, Hinton, and
  Williams]{rumelhart1988learning}
David~E Rumelhart, Geoffrey~E Hinton, and Ronald~J Williams.
\newblock Learning representations by back-propagating errors.
\newblock \emph{Cognitive modeling}, 5\penalty0 (3):\penalty0 1, 1988.

\bibitem[Salimans et~al.(2014)Salimans, Kingma, and
  Welling]{salimans2014markov}
Tim Salimans, Diederik~P Kingma, and Max Welling.
\newblock Markov chain monte carlo and variational inference: Bridging the gap.
\newblock \emph{arXiv preprint arXiv:1410.6460}, 2014.

\bibitem[Serdyuk et~al.(2017)Serdyuk, Ke, Sordoni, Pal, and
  Bengio]{serdyuk2017twin}
Dmitriy Serdyuk, Rosemary~Nan Ke, Alessandro Sordoni, Chris Pal, and Yoshua
  Bengio.
\newblock Twin networks: Using the future as a regularizer.
\newblock \emph{arXiv preprint arXiv:1708.06742}, 2017.

\bibitem[Sordoni et~al.(2017)Sordoni, GOYAL, Cote, Ke, and Bengio]{Sordoni2017}
Alessandro Sordoni, Anirudh Goyal ALIAS~PARTH GOYAL, Marc-Alexandre Cote, Nan
  Ke, and Yoshua Bengio.
\newblock Z-forcing: Training stochastic recurrent networks.
\newblock In \emph{Advances in Neural Information Processing Systems}. 2017.
\newblock URL \url{https://nips.cc/Conferences/2017/Schedule?showEvent=9439}.

\bibitem[Sotelo et~al.(2017)Sotelo, Mehri, Kumar, Santos, Kastner, Courville,
  and Bengio]{sotelo2017char2wav}
Jose Sotelo, Soroush Mehri, Kundan Kumar, Joao~Felipe Santos, Kyle Kastner,
  Aaron Courville, and Yoshua Bengio.
\newblock Char2wav: End-to-end speech synthesis.
\newblock 2017.

\bibitem[Srivastava et~al.(2014)Srivastava, Hinton, Krizhevsky, Sutskever, and
  Salakhutdinov]{srivastava2014dropout}
Nitish Srivastava, Geoffrey~E Hinton, Alex Krizhevsky, Ilya Sutskever, and
  Ruslan Salakhutdinov.
\newblock Dropout: a simple way to prevent neural networks from overfitting.
\newblock \emph{Journal of machine learning research}, 15\penalty0
  (1):\penalty0 1929--1958, 2014.

\bibitem[{Theano Development Team}(2016)]{2016arXiv160502688short}
{Theano Development Team}.
\newblock {Theano: A {Python} framework for fast computation of mathematical
  expressions}.
\newblock \emph{arXiv e-prints}, abs/1605.02688, May 2016.
\newblock URL \url{http://arxiv.org/abs/1605.02688}.

\bibitem[Uria et~al.(2016)Uria, C{\^o}t{\'e}, Gregor, Murray, and
  Larochelle]{uria2016neural}
Benigno Uria, Marc-Alexandre C{\^o}t{\'e}, Karol Gregor, Iain Murray, and Hugo
  Larochelle.
\newblock Neural autoregressive distribution estimation.
\newblock \emph{Journal of Machine Learning Research}, 17\penalty0
  (205):\penalty0 1--37, 2016.

\bibitem[Wang et~al.(2017)Wang, Skerry-Ryan, Stanton, Wu, Weiss, Jaitly, Yang,
  Xiao, Chen, Bengio, et~al.]{wang2017tacotron}
Yuxuan Wang, RJ~Skerry-Ryan, Daisy Stanton, Yonghui Wu, Ron~J Weiss, Navdeep
  Jaitly, Zongheng Yang, Ying Xiao, Zhifeng Chen, Samy Bengio, et~al.
\newblock Tacotron: Towards end-to-end speech syn.
\newblock \emph{arXiv preprint arXiv:1703.10135}, 2017.

\bibitem[Zaremba et~al.(2014)Zaremba, Sutskever, and
  Vinyals]{zaremba2014recurrent}
Wojciech Zaremba, Ilya Sutskever, and Oriol Vinyals.
\newblock Recurrent neural network regularization.
\newblock \emph{arXiv preprint arXiv:1409.2329}, 2014.

\end{thebibliography}
\bibliographystyle{iclr2018_conference}

\newpage
\section*{Appendix}
{\bf{A:}}  Derivation of variation lower bound ${\mathcal{L}}_{\boldsymbol{\Gamma}}$ in equation \eqref{eq:elbow} in more details:
\begin{equation*}
\begin{split}
   \log p(\*x; \boldsymbol{\Gamma}) &=\log\Big[ \prod_{t=0}^T\int_{\*z_{t}}\int_{\tilde{\*b}_t} \Big[p_{\eta}(\*x_{t+1}| \*x_{1:t}, \*z_{t}, \tilde{\*b}_t) p_{\psi}(\tilde{\*b}_t|\*z_{t})  p_{\theta}(\*z_{t} | \*x_{1:t})\Big]d\tilde{\*b}_t  d\*z_{t}\Big]\\
   &=\sum_{t=0}^T\log\Big[ \int_{\*z_{t}}p_{\theta}(\*z_{t} | \*x_{1:t})\int_{\tilde{\*b}_t} \Big[p_{\eta}(\*x_{t+1}| \*x_{1:t}, \*z_{t}, \tilde{\*b}_t) p_{\psi}(\tilde{\*b}_t|\*z_{t})  \Big]d\tilde{\*b}_t  d\*z_{t}\Big]\\
      &=\sum_{t=0}^T\log\Big[ \int_{\*z_{t}}q_\phi(\*z_{t}|\*x_{1:t}) \frac{p_{\theta}(\*z_{t} | \*x_{1:t})}{q_\phi(\*z_{t}|\*x_{1:t})}\int_{\tilde{\*b}_t} \Big[p_{\eta}(\*x_{t+1}| \*x_{1:t}, \*z_{t}, \tilde{\*b}_t) p_{\psi}(\tilde{\*b}_t|\*z_{t})  \Big]d\tilde{\*b}_t  d\*z_{t}\Big]\\
          &\geq\sum_{t=0}^T\int_{\*z_{t}}q_\phi(\*z_{t}|\*x_{1:t})\log\Big[ \frac{p_{\theta}(\*z_{t} | \*x_{1:t})}{q_\phi(\*z_{t}|\*x_{1:t})}\int_{\tilde{\*b}_t} \Big[p_{\eta}(\*x_{t+1}| \*x_{1:t}, \*z_{t}, \tilde{\*b}_t) p_{\psi}(\tilde{\*b}_t|\*z_{t})  \Big]d\tilde{\*b}_t  d\*z_{t}\Big]\\
        &=\sum_{t=0}^T\int_{\*z_{t}}\Big[q_\phi(\*z_{t}|\*x_{1:t})\log(  \frac{p_{\theta}(\*z_{t} | \*x_{1:t})}{q_\phi(\*z_{t}|\*x_{1:t})})  \\
        &\quad\quad + q_\phi(\*z_{t}|\*x_{1:t}) \log\Big[\int_{\tilde{\*b}_t} \Big[p_{\eta}(\*x_{t+1}| \*x_{1:t}, \*z_{t}, \tilde{\*b}_t) p_{\psi}(\tilde{\*b}_t|\*z_{t})  \Big]d\tilde{\*b}_t  d\*z_{t}\Big]\\
     &=\sum_{t=0}^T\Big[\int_{\*z_{t}}q_\phi(\*z_{t}|\*x_{1:t})\log\Big[\int_{\tilde{\*b}_t} \Big[p_{\eta}(\*x_{t+1}| \*x_{1:t}, \*z_{t}, \tilde{\*b}_t) p_{\psi}(\tilde{\*b}_t|\*z_{t})  \Big]d\tilde{\*b}_t  d\*z_{t}\Big]\\
     &\quad\quad- D_{KL}(q_\phi(\*z_{t}|\*x_{1:t})\| p_{\theta}(\*z_{t} | \*x_{1:t}))\Big]\\
     &\geq\sum_{t=0}^T\Big[\int_{\*z_{t}}q_\phi(\*z_{t}|\*x_{1:t})\int_{\tilde{\*b}_t} p_{\psi}(\tilde{\*b}_t|\*z_{t}) \log\Big[p_{\eta}(\*x_{t+1}| \*x_{1:t}, \*z_{t}, \tilde{\*b}_t)  \Big]d\tilde{\*b}_t  d\*z_{t}\\
     &\quad\quad- D_{KL}(q_\phi(\*z_{t}|\*x_{1:t})\| p_{\theta}(\*z_{t} | \*x_{1:t}))\Big]\\
  &\approx  \sum_{t=0}^T\underset{{\*z_{t}\sim  q_\phi(\*z_t | \*x_{1:t})}}{\mathbb{E}}\underset{{\tilde{\*b}_t\sim p_{\psi}(\tilde{\*b}_t|\*z_t)}}{\mathbb{E}}\Big[\log p_{\eta}(\*x_{t+1}| \*x_{1:t}, \*z_{t}, \tilde{\*b}_t) \Big] 
     - D_{KL}(q_\phi(\*z_{t}|\*x_{1:t})\| p_{\theta}(\*z_{t} | \*x_{1:t})).
\end{split}
\end{equation*}
\end{document}